\documentclass[sigconf]{acmart}
\setcopyright{acmcopyright}
\copyrightyear{2018}
\acmYear{2018}
\acmDOI{XXXXXXX.XXXXXXX}

\acmConference[Conference acronym 'XX]{Make sure to enter the correct
  conference title from your rights}{June 03--05,
  2018}{Woodstock, NY}

\acmBooktitle{Woodstock '18: ACM Symposium on Neural Gaze Detection,
 June 03--05, 2018, Woodstock, NY} 
\acmPrice{15.00}
\acmISBN{978-1-4503-XXXX-X/18/06}

\begin{document}

\title{Making Small Language Models Better Multi-task Learners with Mixture-of-Task-Adapters}


\author{Yukang Xie$^{1,2}$, Chengyu Wang$^2$, Junbing Yan$^{3,2}$, Jiyong Zhou$^{1,2}$, Feiqi Deng$^{1}$, Jun Huang$^{2}$}
\affiliation{%
\institution{$^1$ South China University of Technology, Guangzhou, China\\
$^2$ Alibaba Group, Hangzhou, China $^3$ East China Normal University, Shanghai, China}
\country{}
%  \streetaddress{1 Th{\o}rv{\"a}ld Circle}
}
\begin{abstract}
Recently, Large Language Models (LLMs) have achieved amazing zero-shot learning performance over a variety of Natural Language Processing (NLP) tasks, especially for text generative tasks. Yet, the large size of LLMs often leads to the high computational cost of model training and online deployment. In our work, we present ALTER, a system that effectively builds the multi-t\underline{A}sk \underline{L}earners with mix\underline{T}ure-of-task-adapt\underline{ER}s upon small language models (with <1B parameters) to address multiple NLP tasks simultaneously, capturing the commonalities and differences between tasks, in order to support domain-specific applications. Specifically, in ALTER, we propose the Mixture-of-Task-Adapters (MTA) module as an extension to the transformer architecture for the underlying model to capture the intra-task and inter-task knowledge. A two-stage training method is further proposed to optimize the collaboration between adapters at a small computational cost. Experimental results over a mixture of NLP tasks show that our proposed MTA architecture and  the two-stage 
 training method achieve good performance.
Based on ALTER, we have also produced MTA-equipped language models for various domains.
\footnote{All datasets are publicly available. The source codes and model checkpoints are released in EasyNLP~\cite{DBLP:conf/emnlp/WangQZLLWWHL22}. URL:~\url{https://github.com/alibaba/EasyNLP/tree/master/examples/mta}}
\end{abstract}

\begin{CCSXML}
<ccs2012>
   <concept>
       <concept_id>10010147.10010178.10010179.10010182</concept_id>
       <concept_desc>Computing methodologies~Natural language generation</concept_desc>
       <concept_significance>500</concept_significance>
       </concept>
 </ccs2012>
\end{CCSXML}

\ccsdesc[500]{Computing methodologies~Natural language generation}

\keywords{multi-task learning, language model, text generation}


\maketitle

\section{Introduction}

% The rapid emergence of Large Language Models (LLMs) has brought significant changes to the field of Natural Language Processing (NLP)~\cite{llm1,llm2}.
The rapid emergence of Large Language Models (LLMs) has brought significant changes to the field of Natural Language Processing (NLP).
In particular, LLMs (such as ChatGPT\footnote{https://chat.openai.com/} with 175B parameters) have demonstrated powerful abilities to interact with users and solve various NLP tasks in the zero-shot learning setting, whose performance even approaches or exceeds humans in some tasks~\cite{DBLP:journals/corr/abs-2304-01852}.
Yet, the impressive performance of LLMs does not
cover up their potential drawbacks in two aspects.
i) The extremely large parameter size of LLMs
leads to unaffordable computational and usage costs, especially when they need to be fine-tuned or deployed privately for specific applications or in resource-constrained environments.
ii) The decoder-only architecture makes them difficult to maintain high performance over traditional NLP tasks with a highly limited output space such as text classification~\cite{DBLP:journals/corr/abs-2302-06476}. 
% ii) The decoder-only architecture makes them difficult to maintain high performance over traditional NLP tasks with a highly limited output space such as text classification~\cite{DBLP:journals/corr/abs-2302-10724,DBLP:journals/corr/abs-2302-06476}. 

Based on the observation,
we revisit the exploitation of~\emph{small} language models for multi-task learning across a variety of NLP tasks, which are much easier to train and deploy.\footnote{By~\emph{small} language models in our work, we refer to language models with <1B parameters, compared to ChatGPT-like models with over 100B parameters.}
Yet, improving the multi-task solving capacities of such models has been non-trivial due to the highly limited parameter space.
Previously, several works have been conducted to improve the multi-task learning abilities of the transformer architecture.
To begin with, Mixture-of-Experts (MoE)~\cite{moe} divides complex tasks into smaller, more manageable sub-problems, each of which is solved by an expert model. Multi-task MoE (MMoE)~\cite{ma2018modeling} uses a multi-gating module to generate specific expert weights according to different tasks.
Switch Transformers~\cite{st} introduces a sparse addition of MoE to solve the problem of large computational efforts. 
However, the above MoE-based works are mainly proposed to train large language models at scale by adding parallel modules and partitioning the data at the token level.
Hence, there is a lack of task-level differentiation for multi-task model adaption. 
In addition, the collaborative relationships between the internal modules (i.e, experts) for better multi-task learning performance are insufficiently explored.

In our work, we introduce the ALTER system that can effectively extend the multi-task learning capacity of~\emph{small} language models to address multiple NLP tasks simultaneously.
In ALTER, the Mixture-of-Task-Adapters (MTA) architecture is designed as a lightweight extension to the transformer architecture in order to capture the commonalities and differences between tasks. A two-stage training method is further introduce to slightly adjust the collaboration between adapters for multiple tasks at a small computational cost. 
We conduct a series of experiments over a mixture of NLP tasks and show that our proposed MTA approach achieves good performance with few additional training costs, compared to standard supervised fine-tuning.
We further release several domain-specific MTA-equipped language models and showcase their values to support real-world applications.

% 模型架构图怎么画才比较合理
\begin{figure*}[ht]
\centering
\includegraphics[width=0.975\textwidth, keepaspectratio]{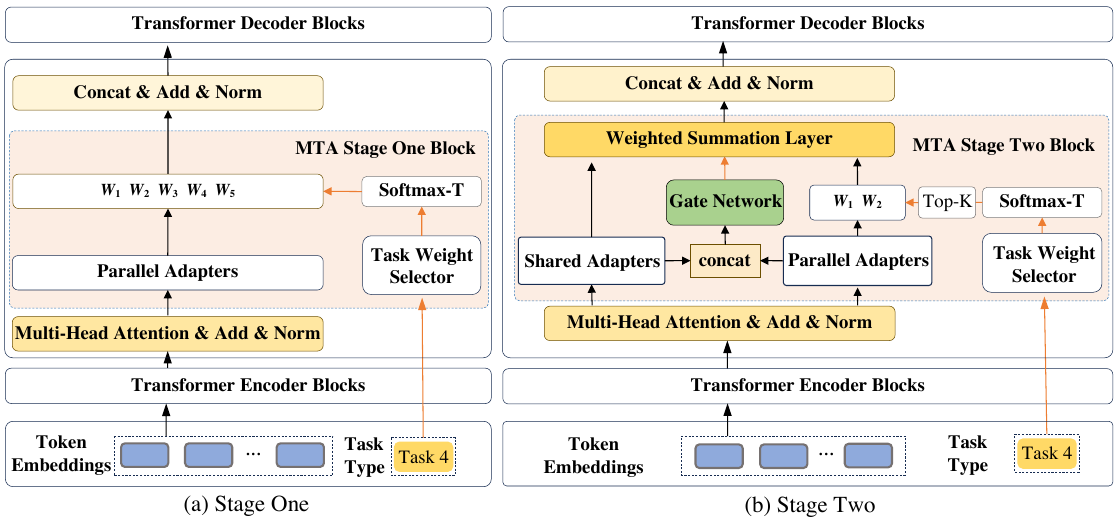}
\caption{The replacement of the FFN (Feed Forward Network) module in a specific transformer layer with the MTA (Mixture-of-Task-Adapters) module, the structure of the adapter mirrors the FFN structure.}
\label{case}
\end{figure*}

\section{The ALTER System}

In this section, we first introduce our system pipeline. After that, we describe the structure of our MTA architecture and how to learn the parameters of MTA on a multi-task dataset. Specifically, the model is trained in two stages, with the parameters of the whole model fine-tuned in the first stage. In the second stage, we add the shared adapter module to improve the model performance by 
% training only shared adapters and the gating network. 
training only the parameters of the MTA module with other parameters frozen.

\begin{figure}
\centering
\includegraphics[width=0.5\textwidth, keepaspectratio]{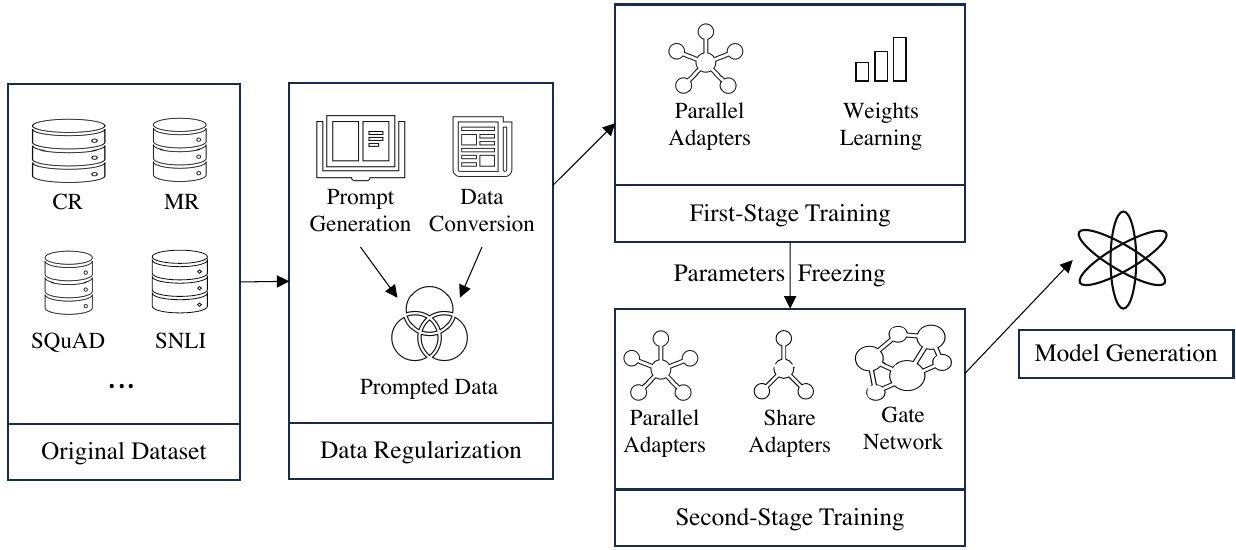}
\caption{The system pipeline of ALTER.}
\label{pipe}
\vspace{-.5em} 
\end{figure}

%The ALTER system pipeline is shown in Figure~\ref{pipe}. As seen, given a multi-task training set of any arbitrary domain, a generative pre-trained language model and the detailed training settings, ALTER automatically converts the model checkpoint to fit the MTA architecture by replacing the FFN (Feed Forward Network) of given transformer layers into MTA. After that, a two-stage training process is called in a sequential order. When this process converges, users can export the modified model checkpoint and deploy it to the corresponding applications. For ease of use, we provide simple and easy-to-use WebUI interfaces that allow users to interact with the model by conversations.
%修改了描述

The system pipeline is shown in Figure~\ref{pipe}.
Given a collection of datasets of any arbitrary domain (possibly including text classification, language inference, generation, etc.),
we regularize all datasets by reformatting the data and adding corresponding prompts, thus obtaining a multitasking dataset that is uniformly formatted. The data are trained in one stage with parallel adapters and weight learning to minimize the interference of different tasks, and then parameter freezing, introduction of shared adapters and gate\cite{ma2018modeling} networks to further improve the collaboration between different tasks. We will subsequently elaborate on the details.

\subsection{First-Stage Training for Obtaining Task-to-Adapter Correspondence}

% The difficulty in improving the performance of multi-task learning lies in how to coordinate the collaboration and interference between different tasks.
% In the first stage, we abandon the token-level input format in MoE~\cite{st} and instead favor the sentence-level input. In our work, we propose the MTA (Mixture-of-Task-Adapters) architecture, which can be seen as parallel adapter blocks. As shown in Figure \ref{case}, in the first stage, the MTA module mainly consists of Parallel Adapters and Task Weights Selector.
In the first stage of training, our main goal is to learn the correspondence between the adapters and tasks.  We abandon the token-level input format in MoE~\cite{st} and instead favor the sentence-level input, the entire sentence is taken as a whole and is not broken down into individual tokens to be entered into a module. In our work, we propose the MTA (Mixture-of-Task-Adapters) architecture, which can be seen as parallel adapter blocks. As shown in Figure \ref{case}, in the first stage, the MTA module mainly consists of Parallel Adapters and Task Weights Selector.

% 这一部分描述不是很准确，但是不知道怎么改
% It can be seen from the figure that the MTA Block replaces the FFN structure in the original transformer block and is lightly embedded.

% To establish a correspondence between tasks and adapters, whereby each adapter is dedicated to a specific type of task, we introduce a bias to the initialization of the weights in the Task Weight Selector. By doing so, we ensure that the model is capable of enabling task-to-adapter correspondence. To amplify the effect of the bias, Softmax-T is employed, thereby enhancing the model's adaptation capability.

\begin{figure}
\centering
\includegraphics[width=0.45\textwidth]{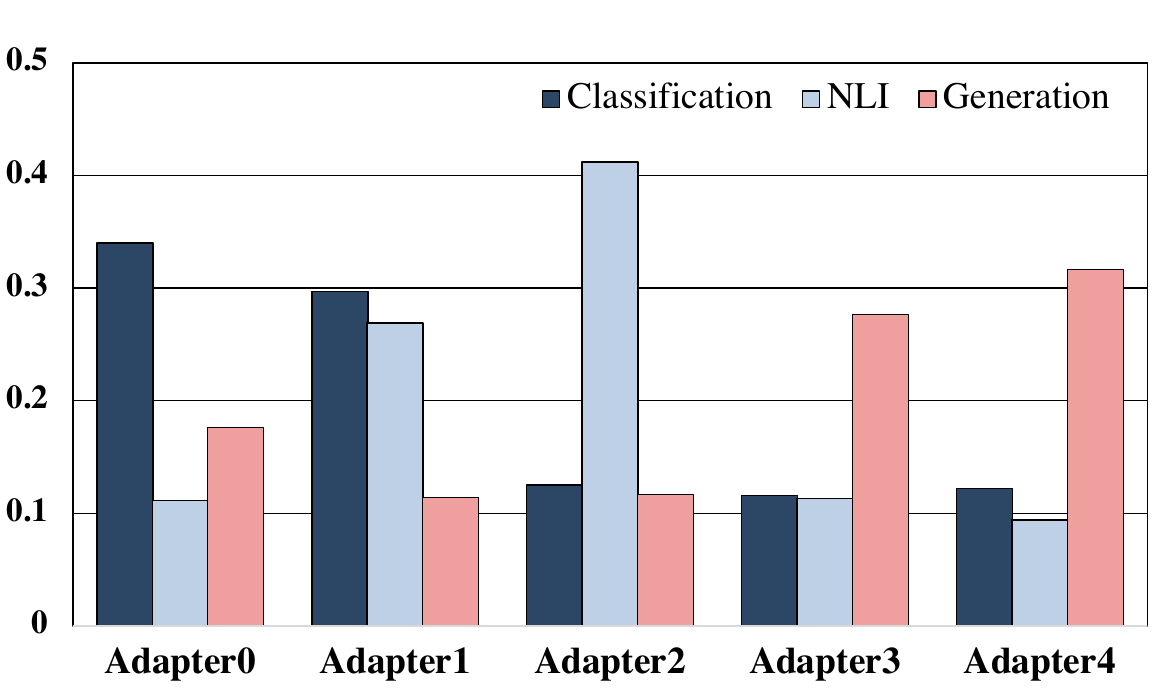}
\caption{Visualization of adaptive weights obtained after first-stage training reveals that weights are strongly correlated with the type of task.}
\label{weight}
\vspace{-.5em}
\end{figure}

To establish a correspondence between tasks and adapters, with each adapter being dedicated to a specific type of task, we introduce a bias to the initialization of the weights in the Task Weight Selector. This ensures that the model is capable of enabling task-to-adapter correspondence. Using Softmax-T amplifies the effect of the bias and enhances the model’s adaptation capability.
For example, the weights of the type of $i$-th task $W_i$ can be initialized as follows:
\begin{equation}
    W_i=[\frac{1}{N}, \frac{1}{N}, \frac{1+\lambda}{N}, \cdots, \frac{1}{N}]
\end{equation}
where $N$ denotes the number of parallel adapters, $\frac{1+\lambda}{N}$ refers to the third gating value, guiding the third adapter to pay more attention to the $i$-th type of task.
Note that during model training, these weights are also learnable, making our mechanism self-adaptive.
%e.g., we initialize the weights of the five adapters to [0.22,0.22,0.20,0.20,0.20] to guide the first two adapters to learn mainly this type of task.
% 熊
% To ensure that the model can achieve task-to-adapter
% correspondence where each adapter focuses on a specific type of tasks to a certain extent,
% we further introduce a number of adaptive weights as gating blocks. Formally, we have:
% 熊
The MTA module during the first phase of training has the following formula:
\begin{equation}
    A{(x)}=Concat(A_1(x),...,A_N(x))
\end{equation}
\begin{equation}
    MTA_{out1}=softmax(\frac{W}{T})\cdot A(x)
\end{equation}
where 
$A_i(x)$ is the output of the $i$-th adapter network w.r.t. the input $x$.
%the input to the hidden layer.
$W$ denotes the manually initialized matrix of gates for all the types of tasks. \textit{T} is the sharpening coefficient. 
%$A$ denotes Adapters, x is input to Adapters to get the output of each Adapter, and then the sharpened weights are weighted and summed to get the output of the MTA module, with exactly the same input and output dimensions.
% 熊
% After initializing the adaptive weights in gates, a 1\% bias is added at the corresponding position of the two adapters to guide the model in multi-task learning, i.e.,
% \begin{equation}
%     a+b=c
% \end{equation}
% 熊
By introducing the gate values, we guide each adapter to make major contributions to one type of tasks only.
%to make major contributions to specific few adapters for a particular type of task, while other adapters made minor contributions. 
This approach enables us to retain the specific characteristics of individual tasks. Take our experimental multi-task datasets as an example. The self-adaptive weights learned during the first stage are shown in Figure \ref{weight}.
In addition, we insert a special token ``[START]'' to each input sample, aiming to capture the semantics of the whole input sequence.

% \begin{figure}[ht]
% \centering
% \includegraphics[width=0.4\textwidth]{weight_6.12.pdf}
% \caption{Visualization of adaptive weights obtained after first-stage training reveals that weights are strongly correlated with the type of task. Taking text classification as an example, Adapter0 and Adapter1 make major contributions, while the remaining adapters make minor contributions.}
% \label{weight}
% \end{figure}

\subsection{Second-Stage Training for Adjustment of Collaboration Relationships}
After completing the first stage of training, we inherit all the weights from the first stage.
% Specifically, we only select the results of top-$K$ gates as the input for the second stage.
To further adjust the collaborative relationship between adapters, we introduce additional shared adapters and a gate network, while freezing all training parameters except for the MTA module. The gate network is used to generate adaptive weights, aiming to obtain collaborative relationships among various adapters for different tasks by capturing the ``[START]'' information in the hidden layer. The formula for the gate network is as follows: 
\begin{equation}
    A^*(x)=concat(\sum_{i=1}^{K}A_i(x)\cdot W_i,S(x))
\label{function2}
\end{equation}
\begin{equation}
    W^*=G(concat(S(x)_{[START]},A^*(x)_{[START]}))
\label{function3}
\end{equation}
\begin{equation}
    MTA_{out2}=A^*(x)\cdot W^*
\label{function1}
\end{equation}
where $A^*(x)$ denotes the combined representations of shared and top-$K$ adapter modules. 
%$W^K$ and $A^K$ denote the result of TopK weights and its corresponding adapters, respectively. 
$S(x)$ is the output of the shared adapters.
%$W$ inherit the final weights of the stage one.
$G$ denotes a gate network with linear layers and activation functions.

%The addition of shared adapters provides an interaction module based on the completion of the first-stage training. Original adapters may not have achieved the optimal collaborative relationship between tasks and can be further trained to target the relationship between adapters through the addition of shared adapters and parameter freezing in the second stage. Thereby, we can reduce the interference between tasks and enhance the collaborative effect between tasks.
Relying only on the first stage parallel adapters leads to direct interactions between tasks and makes it difficult to achieve a better state of collaboration. 
% In the second stage, this overly direct interaction can be solved by adding shared adapters Top-k selecting and freezing parameters to convey the collaboration between tasks through shared adapters. Therefore, we can reduce the direct interference between tasks and further enhance the collaboration between tasks.
In the second stage, the top-k selection can eliminate direct interference between different tasks. To compensate for the collaborative effect of synchronous deletion without reintroducing direct interference, a shared adapter structure is added as a transition module for information interaction to moderate the direct interference between tasks.

% \subsection{Training}
% Our model is trained in a two-stage training approach. The first stage is the overall training, in which all parameters of the model are fine-tuned to obtain a model that can generalize the dataset well, and the second stage is the partial fine-tuning, in which shared adapters are added to the MTA and then only the parameters of the MTA module and shared adapters are updated. The second stage of fine-tuning is to further optimize the co-operative approach and the parameters in the adapters to further improve the model performance.

\section{Experimental Evaluation}

In this section, we evaluate the effectiveness of our proposed approach over public datasets.

\subsection{Datasets and Experimental Settings}

We employ three common types of NLP tasks (text classification, natural language inference task, and question answering) to construct multi-task datasets. The datasets include:
\begin{itemize}
    \item Text classification: CR~\cite{crdataset}, MR~\cite{mrdataset}, SST-2~\cite{sst-2} and TREC~\cite{trecdataset};
    \item Natural language inference: SNLI~\cite{snli};
    \item Question answering (text generation): SQuAD~\cite{squaddataset}.
\end{itemize}
%and summarized several common datasets including classification task dataset (cr,mr,SST-2,trec), natural language inference task dataset (SNLI), and generation task dataset (squad). 

Following the work of T5 \cite{t5}, we add appropriate prompts for all the tasks to convert them into a uniform input format.
For classification and inference tasks, we match the model outputs with the ground-truth and report the prediction accuracy. For generation, we constructed the data in such a way as to generate questions based on the answers, we report the Rouge-L score \cite{rouge}, measuring the similarity between generated and reference texts.
In the implementation, we load the pre-trained weights from T5-base, with 24 transformer layers.
Baseline models include the base versions of BART \cite{lewis2019bart}, GPT-2 \cite{GPT-2} and Switch Transformer \cite{st}.
The performance of ChatGPT is also reported for reference.
For our method, we also conduct comparative experiments on T5-large. %All the models are trained on a server with 4 NVIDIA V100 (32G) GPUs.

%We compare the same size T5-Base, ST and our method, both using a 24 layers transformer architecture (12 layers encoder, 12 layers decoder), train 20 epochs on our dataset and saving one optimal model per epoch, We also conducted comparative experiments on T5-Large. We complete the training on 8 NVIDIA 32G V100 GPUs. Moreover, we selected several commonly used models for multitasking, including ChatPGT, Flan-t5, GPT-2, and BART, to evaluate the dataset we constructed.

\begin{table}
\centering
\begin{tabular}{ccccc}
\hline
\textbf{Method} & \textbf{Classification} & \textbf{NLI} & \textbf{QA} & \textbf{Score} \\
\hline
ChatGPT  & 35.88 &  43.90 & 72.15 & 50.70\\
\hline
BART  & 88.23 & 78.00 & 72.92 & 79.72\\
GPT-2  & 91.58 & 79.63 & 71.88 & 81.03\\
Switch Trans. & 91.10 & 82.85 & 62.70 & 78.88\\
\hline
T5-Base  & 92.47 & 84.78 & 80.00 & 0.8575\\
Ours-Base  & \textbf{93.26} & \textbf{86.34} & \textbf{81.28} & \textbf{86.96}\\
\hline
T5-Large  & 93.81 & 88.54 & 77.82 & 86.72\\
Ours-Large  & \textbf{94.01} & \textbf{88.71} & \textbf{82.70} & \textbf{88.47}\\
\hline
\end{tabular}
\caption{Overall evaluation results. ChatGPT is evaluated under the zero-shot learning setting and is for reference only. NLI and QA are short for natural language inference and question answering, respectively.}
\label{maintable}
\vspace{-1em}
\end{table}

\begin{table}
\centering
\begin{tabular}{ccc}
\hline
\textbf{Model} & \textbf{Score} (T5-base) & \textbf{Score} (T5-large)\\
\hline
\bf Full implement. & \textbf{86.96} & \textbf{88.47}\\
\hline
Vanilla fine-tuning & 85.75 & 86.72\\
w/o. second-stage & 86.58 & 87.81\\ 
w/o. parameter freeze & 86.60 & 87.40 \\
\hline
\end{tabular}
\caption{Ablation results of two-stage training with different strategies on T5-base and T5-large models.}
\label{tab:strategies}
\vspace{-2em}
\end{table}

\subsection{General Experimental Results}
In the first set of experiments,
%we set up, 
we mix multiple tasks into a test set, and then obtain four scores via a unified scoring criterion.
%which contained the final scores and scores for the three types of tasks. 
Several points can be observed from Table \ref{maintable}. Firstly, our model achieves higher overall results than other models, demonstrating the effectiveness of our proposed method. Secondly, when compared to the token-level processing method of Switch Transformer, our sentence-level approach to task processing exhibits a significant performance advantage on this multi-task dataset. Finally, we compare the results of our proposed model with ChatGPT-turbo. The outputs generated by ChatGPT-turbo are more difficult to control due to its intrinsic properties. Therefore, we relax the scoring criteria to include answer words as long as they are correct for both classification tasks and natural language inference tasks. Despite this relaxation, our model still outperforms the ChatGPT-turbo baseline, demonstrating its effectiveness even against larger models after fine-tuning on common multi-task datasets.

% \begin{figure}[ht]
% \centering
% \includegraphics[width=0.46\textwidth]{dataset_result.pdf}
% \caption{The figure displays the experimental results of multi-task training on four different datasets of varying task ratios.}
% \label{tab:dataset}
% \end{figure}

% As shown in the table \ref{tab:num_adapters}, Adding more adapters can improve model performance when the number of adapters is low, but it can lead to a decrease in performance when the number of adapters exceeds 4.

% \begin{table}
% %\vspace{-1.5em}
% \centering
% \begin{tabular}{cccc}
% \hline
% \textbf{Proportion} & \textbf{Base} & \textbf{Stage 1} & \textbf{Stage 2} \\
% \hline
% 1:1:1  & 86.20 &  +0.53\% & \textbf{+0.92\%} \\
% 1:2:2  & 85.47 &  +0.76\% & \textbf{+0.87\%} \\
% 2:1:2  & 85.90 &  +0.49\% & \textbf{+0.84\%} \\
% 2:2:1  & 85.75 &  +1.04\% & \textbf{+1.28\%} \\
% \hline
% \end{tabular}
% \caption{Result analysis when we use four different proportional mixes of datasets.
% %for the two-stage training results, and 
% The results of Stage 1 and Stage 2 are marked with base-based improvements.}
% \label{tab:dataset}
% \vspace{-1.0em}
% \end{table}

\begin{figure}
\centering
\includegraphics[width=0.48\textwidth]{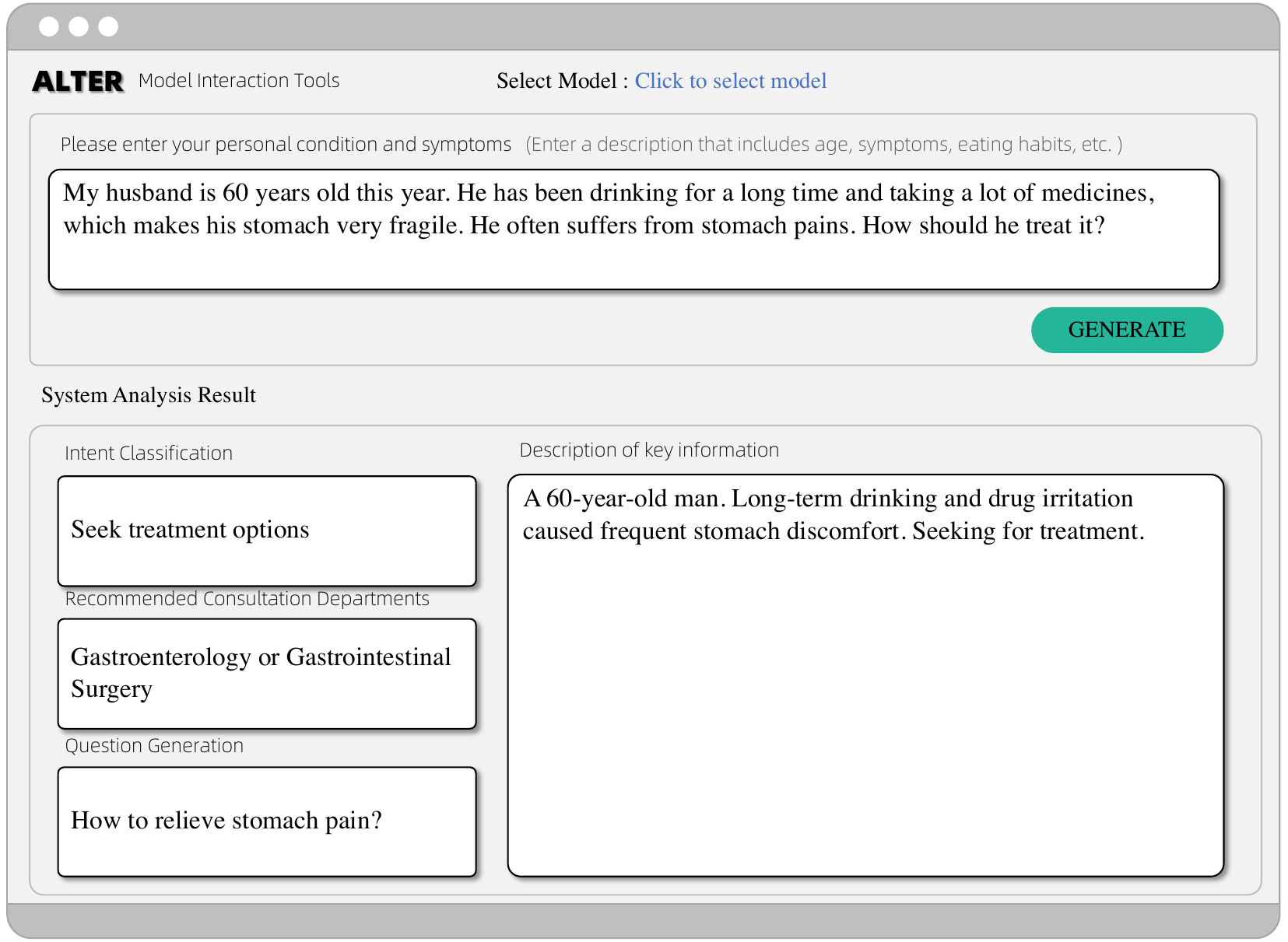}
% \caption{The inference process and results of the model trained on the PromptCBLUE dataset presented on WebUI. Questions and answers are originally in Chinese and have been translated into English for reference.}
\caption{The inference process and results of the model trained on the PromptCBLUE dataset presented on WebUI.}
\label{fig:webui}
\vspace{-1em}
\end{figure}

\subsection{Ablation Study and Model Analysis}
% To further demonstrate the effectiveness of our proposed two-stage training method and the shared adapter, we conduct a series of ablation experiments.
To further demonstrate the effectiveness of our proposed two-stage training method and the shared adapter, we conduct a series of ablation experiments. 
Due to space limitations, we show only the most representative set of ablation experiments.
As shown in Table \ref{tab:strategies}, models with multiple adapter structures show significant improvements compared to the original model, regardless of whether we use T5-base or T5-large. 

Our experiments have revealed that partial parameter fine-tuning performs better and requires less computational resources than full fine-tuning during the second stage of training. Furthermore, a comparison with the training results of the two-stage model architecture using global parameter fine-tuning shows that partial fine-tuning can optimize the collaboration between adapters and significantly improve model performance.

% Further comparison with the results obtained by training a two-stage model architecture from scratch demonstrates that partial fine-tuning can optimize the collaboration between adapters and substantially improve the model performance. 

% As our experiments involve datasets of three general NLP tasks: text classification, natural language inference and text generation, we re-sample the training sets and use four different proportional mixes of datasets for training. 
% Our results in Table \ref{tab:dataset} demonstrate that our proposed method remains effective under various task distributions, showing its generalization abilities for different tasks.

% \section{Demonstration Scenarios}
% In this demo, we will show the full process of using the model we have trained. A snapshot of the results comparing the performance of our trained model and the larger model on the same task can be found in Figure \ref{fig:webui} below. It can be seen that our model can achieve similar experimental results as the large model, and also for domain-specific applications, we have trained on a specific task category of the PromptCBLUE medical dataset (Chinese dataset), and the results of comparing the large model are shown in Figure 2 below.

\section{Demonstration Scenarios}
In this demo, we will show the complete process of how our models are trained and deployed. 
%the full process of using the model we have trained.
%A snapshot of the results comparing the performance of our trained model and the larger model on the same task can be found in Table \ref{maintable} above. It can be seen that our model can achieve similar experimental results as the large model, and also 
For domain-specific applications, we have also trained
several MTA-equipped language models for various domains.
Take medicine for an example. Our model is trained over the multi-task PromptCBLUE dataset\footnote{https://github.com/michael-wzhu/PromptCBLUE}, which is a large-scale instruction-tuning dataset in the medical domain in Chinese.
%(Chinese dataset) and selected 
The results also prove that our generated model, despite in small parameter size, can generate domain-specific knowledge accurately.
For better human-machine interactions, we have set up  the WebUI interface, as shown in Figure \ref{fig:webui}.

\section{Conclusion}

We propose ALTER, a system that effectively builds the multi-t\underline{A}sk \underline{L}earners with mix\underline{T}ure-of-task-adapt\underline{ER}ts to address multiple NLP tasks simultaneously.
Specifically, the Mixture-of-Task-Adapters (MTA) are introduced to help models learn task differences and commonalities to improve model performance on multi-task datasets. We demonstrate that our proposed method can achieve performance comparable to that of large models.
%on specific multi-task datasets, using our own constructed multi-task dataset. 
In the proposed MTA module, we introduce a two-stage training method that incorporates prior knowledge to obtain basic collaborative model parameters for specific adapters corresponding to specific tasks in the first stage, and further improves model performance by coordinating collaboration among adapters in the second stage.
%Our method has a significant advantage over several common multi-task language models of the same size, with faster two-stage training that leads to better performance improvement compared to global parameter fine-tuning. MTA is a versatile module and exploring the effectiveness of the MTA modules and two-stage training on other model architectures is the focus of our work.
We further demonstrate how to apply our technique for domain-specific applications.


\end{document}